\documentclass[letterpaper, 10 pt, conference]{ieeeconf}  

\IEEEoverridecommandlockouts                              
\usepackage[utf8]{inputenc}
\usepackage{graphicx} 
\usepackage{amsmath} 
\usepackage{amssymb}  
\usepackage{color}

\usepackage{lipsum}
\usepackage[colorinlistoftodos]{todonotes}
\usepackage{multirow}
\usepackage{subcaption}
\usepackage[ruled,vlined]{algorithm2e} 
\usepackage{url}
\usepackage{booktabs}
\usepackage{tabularx}
\usepackage{balance}
\newcolumntype{Y}{>{\centering\arraybackslash}X}
\usepackage{siunitx}
\makeatletter
\newcommand{\thickhline}{%
    \noalign {\ifnum 0=`}\fi \hrule height 1pt
    \futurelet \reserved@a \@xhline
}
\newcolumntype{"}{@{\hskip\tabcolsep\vrule width 1pt\hskip\tabcolsep}}
\makeatother

\usepackage[T1]{fontenc}
\usepackage{textcomp}

\newcommand{\vtr}{{VT\&R}}

\usepackage{amsmath}
\usepackage{amssymb}
\usepackage{amsfonts}
\usepackage{dsfont}
\usepackage{etoolbox}	

\newcommand{\etal}{\emph{et al.}}

\newcommand{\mahalanobisNorm}[2]{\lVert{#1}\rVert^{2}_{#2}}

\newcommand{\argmin}{\operatornamewithlimits{arg\,min}}

\newcommand{\subarrow}[1]{
	\mathord{
		\renewcommand{\arraystretch}{0}
		\begin{array}[t]{@{}c@{}l@{}}
			#1\\[2pt]
			\hspace{-2pt}\scriptstyle\longrightarrow
		\end{array}
		\kern\scriptspace
	}
}
\newcommand{\notatFrame}[1]{\subarrow{\mathcal{F}}{}_{#1}}

\newcommand{\notatMatrix}[1]{\boldsymbol{\mathrm{#1}}}
\newcommand{\notatVector}[1]{\boldsymbol{\mathrm{#1}}}
\newcommand{\notatScalar}[1]{{#1}}
\newcommand{\notatHomog}[1]{\boldsymbol{{#1}}}
\newcommand{\notatManifold}[1]{\mathcal{\MakeUppercase{#1}}}

\newcommand{\notationMatrix}[2]{\boldsymbol{\mathrm{#1}}_{ #2}}
\newcommand{\notationVector}[2]{\boldsymbol{\mathrm{#1}}_{ #2}}
\newcommand{\notationScalar}[2]{{#1}_{#2}}
\newcommand{\notationHomog}[2]{\boldsymbol{{#1}}_{#2}}

\newcommand{\notationMatrixFrame}[3]{{_#2}\boldsymbol{\mathrm{#1}}_{#3}}
\newcommand{\notationVectorFrame}[3]{{_#2} \boldsymbol{\mathrm{#1}}_{#3}}
\newcommand{\notationScalarFrame}[3]{{_#2}{#1}_{#3}}
\newcommand{\notationHomogFrame}[3]{{_#2}\boldsymbol{{#1}}_{#3}}

\robustify{\notatFrame}
\robustify{\notatMatrix}
\robustify{\notatVector}
\robustify{\notatScalar}
\robustify{\notatHomog}
\robustify{\notationMatrix}
\robustify{\notationVector}
\robustify{\notationScalar}
\robustify{\notationHomog}
\robustify{\notationMatrixFrame}
\robustify{\notationVectorFrame}
\robustify{\notationScalarFrame}
\robustify{\notationHomogFrame}

\newcommand{\SEthree}{\mathrm{SE(3)}}

% ...

\makeatletter
\renewcommand*\env@matrix[1][\arraystretch]{%
	\edef\arraystretch{#1}%
	\hskip -\arraycolsep
	\let\@ifnextchar\new@ifnextchar
	\array{*\c@MaxMatrixCols c}}
\makeatother

\usepackage{color}
\usepackage{colortbl}
\definecolor{ColorLightCyan}{rgb}{0.88,1,1}
\definecolor{ColorLightTurquoise}{rgb}{0.5, 1, 0.8}

\def\secref#1{Sec.~\ref{#1}}
\def\figref#1{Fig.~\ref{#1}}
\def\tabref#1{Tab.~\ref{#1}}
\def\eqref#1{Eq.~(\ref{#1})}
\def\algref#1{Alg.~\ref{#1}}

\title{
\LARGE \bf
Learning Camera Performance Models for \\Active Multi-Camera Visual Teach and 
Repeat
}

\author{Matias Mattamala, Milad Ramezani, Marco Camurri, and Maurice Fallon
\thanks{The authors are with the Oxford Robotics Institute at the University of Oxford, UK. 
{\tt\small \{matias, milad, mcamurri, mfallon\} @robots.ox.ac.uk }
}
}

\begin{document}
\bstctlcite{IEEEexample:BSTcontrol}

\maketitle


\begin{abstract}
In dynamic and cramped industrial environments, achieving reliable Visual Teach 
and Repeat (\vtr{}) with a single-camera is challenging. In this work, we 
develop a robust method for non-synchronized multi-camera \vtr{}. Our 
contribution are expected Camera Performance Models (CPM) which evaluate the
camera streams from the teach step to determine the most informative one for 
localization during the repeat step. By actively selecting the most suitable
camera for localization, we are able to successfully complete missions when one 
of the cameras is occluded, faces into feature poor locations or if the 
environment has changed. Furthermore, we explore the specific challenges of
achieving \vtr{} on a dynamic quadruped robot, ANYmal. The camera does 
not follow a linear path (due to the walking gait and holonomicity) such that precise path-following cannot be achieved.
Our experiments feature forward and backward facing stereo cameras 
showing \vtr{} performance in cluttered indoor and outdoor scenarios. 
We compared the trajectories the robot executed during the repeat steps demonstrating typical
tracking precision of
less than \SI{10}{\centi\meter}
on average. With a view towards omni-directional localization, we show how the
approach generalizes to four cameras in simulation.
\end{abstract}



\section{Introduction}
Following previously traversed paths is a useful capability for mobile robots. 
This is essential for missions, such as autonomous inspection and monitoring,
where the same path is repeatedly traversed. This has motivated research into 
mapping and localization systems \cite{Cadena2016a}. In particular,
vision-based navigation systems such as Visual Teach and Repeat 
(\vtr{}) \cite{Furgale2010} have enabled different
robots to repeat known routes without requiring metrically accurate maps. 
Visual sensors are inexpensive, lightweight, and provide both appearance 
and geometric information about the robot's surroundings.

We are interested in legged robots, which are promising for inspection
tasks due to their versatile mobility on challenging terrains. However,
quadrupeds such as ANYmal \cite{Hutter2017} are holonomic and move with dynamic
gaits, such as trotting and climbing stairs. These motions cause rapid
feature change, blur and
tracking failure, making it difficult to achieve
\vtr{} with a single camera. Since cameras have a limited Field-of-View (FoV),
redundancy in visual sensing, i.e using multiple cameras, allows 
such platforms to increase their vision capabilities, but at an increased 
computational cost and integration complexity (synchronization and 
calibration).

\begin{figure}[t]
	\centering
	\includegraphics[width=\columnwidth]{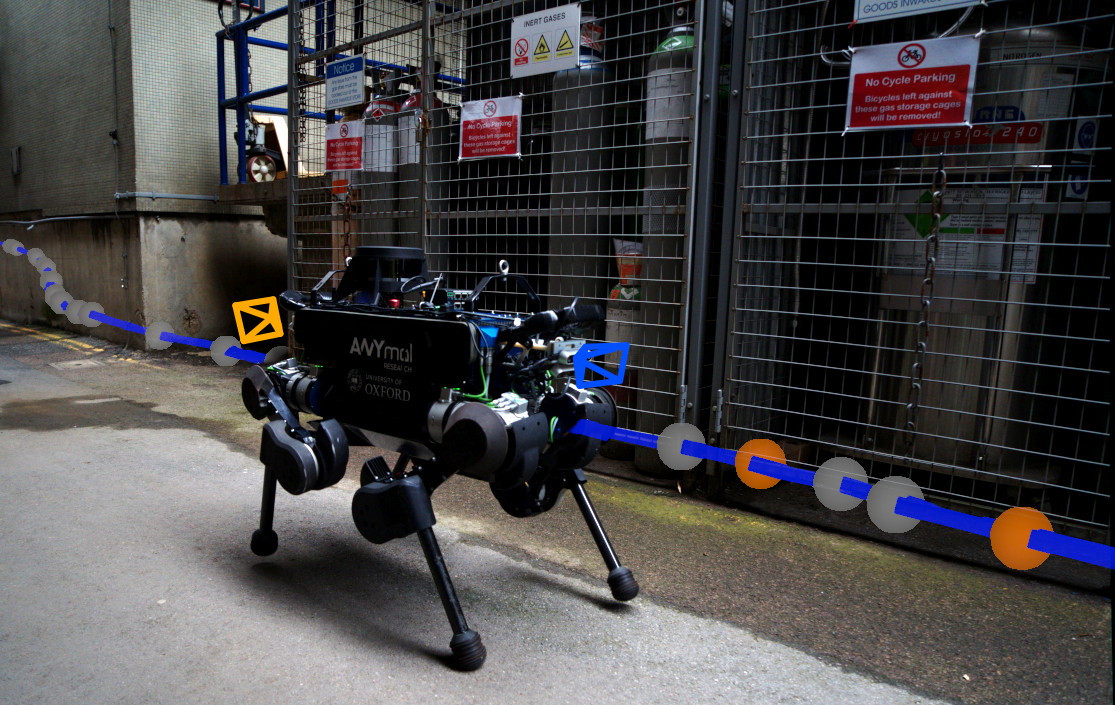}
	\caption{\small{Visual Teach and Repeat (\vtr{}) allowed us to quickly deploy the
		ANYmal robot for industrial routine inspections. In this work, we
extend previous approaches by augmenting the topo-metric map with
		a \emph{Camera Performance Model} (CPM). This allows us to dynamically
		choose the most reliable camera during a repeat, such as the front
		(blue) and rear (orange) cameras illustrated.}}
	\label{fig:anymal-industrial}
	\vspace{-3mm}
\end{figure}

In this work, we present a visual navigation system for mobile robots based on
the \vtr{} paradigm that takes advantage of multiple cameras to stay localized
in presence of clutter in narrow spaces (\figref{fig:anymal-industrial}).
In contrast to previous approaches, which typically process hardware 
synchronized cameras simultaneously~\cite{Forster2017a, Liu2018,Won2020}, we
instead select the camera providing the best performance for each segment of 
the path. This approach allows us to achieve accurate path
tracking while also being robust to dynamic changes in the environment.
Since cameras hardware synchronization is not required, our approach is
more flexible, scalable and easier to deploy than traditional methods.
The contributions of our work are summarized as follows:

\begin{itemize}
 \item A \vtr{} system that uses multiple cameras during the \emph{teach} step
 and learns performance models for each stream. These models are used in the  
 \emph{repeat} step to actively select the most informative camera.
 \item Deployment of our \vtr{} system on a quadruped robot,
 which enables it to autonomously follow a path it has traveled before 
 using only vision, in spite of occlusions and dynamic locomotion gaits.
\item Evaluation of the system in simulated and real scenarios with an ANYmal 
quadruped where peculiarities of legged systems are discussed. To the
best of our knowledge, this is the first academic demonstration of \vtr{} on a
legged platform.
\end{itemize}

The remainder of this paper is structured as follows.~\secref{sec:related-work} 
discusses the related work.~\secref{sec:method} describes our Active
Multi-Camera \vtr{} system. Experimental results are presented
in~\secref{sec:experiments} and conclusions are drawn
in~\secref{sec:conclusions}.

\section{Related Work}
\label{sec:related-work}
This section discusses previous \vtr{} approaches as well as methods that 
exploit mapping or teach steps to improve the performance in subsequent 
traversals. 

\subsection{Visual Teach and Repeat}
A variety of \vtr{} systems have been developed for wheeled robots~\cite{Furgale2010a, 
Krajnik2018} and drones~\cite{Gao2019, Warren2019, Nitsche2020}. The main idea
behind \vtr{} is that a topo-metric feature map is collected during a teach
run. The map can then be used to guide the robot along the path learned during
the teach run. Only local consistency between the path and the map is required
to achieve path following.

In the past 10 years, most research on \vtr{} has been focused on
improving the robustness against long-term environmental changes, which can
compromise visual navigation. The problem has been commonly approached by
creating and analyzing a varied set of traversals of the same route (also
called \emph{experiences}). This approach is known as \emph{Experience-Based
Navigation (EBN)}~\cite{Churchill2012} or
\emph{Multi-Experience Localization (MEL)}~\cite{Paton2016}. Both systems have 
been deployed in autonomous cars, ground, and aerial platforms, with
emphasis on seasonal and day-night reliability~\cite{Linegar2016,
Paton2017, Warren2019}.

In the past, using multiple cameras in the context of \vtr{} has
only been applied to deal with appearance changes. Paton 
\etal~\cite{Paton2015} used front-view and rear-view synchronized cameras on a 
Husky robot to make their \vtr{} system more robust to lighting conditions. 
However, this process was passive, as both cameras were processed together, and 
no prior information about the path was utilized.

In this work, we are less
concerned about long term changes, such as day-night shifts or
seasonal changes. Instead, we focus on abrupt
changes, such as motion blur due to aggressive motions, occlusions
caused by
people or vehicles, camera exposure changes, and rapid scene change in
cluttered locations. To this end, we also explore the use of multiple
cameras, but we actively select the most informative camera during the repeat
step. This is done by comparing online the current and the expected
performance inferred from previous traversals.

\subsection{Leveraging Past Experiences}
Because \vtr{} systems have an explicit teach step, the knowledge collected in 
this step can be used to optimize the performance during the repeat
execution. As with other methods, the main assumption is that the route
taken does not change drastically, so the collected information (features
and models) can be leveraged in future operation.

The work of Churchill \etal~\cite{Churchill2015} collected localization
statistics from several teach passes to train a Gaussian Process (GP) model of 
the \emph{localization envelope} of a given path. This embedded the
localization performance of the system, and it was used to predict potential 
failures.

Ondruska \etal~\cite{Ondruska2015} showed how to reduce the energy requirements 
of a planetary rover by scheduling when to use its cameras. They showed 
that significant energy savings could be achieved while still reliably 
localizing in open, park-like environments. Warren \etal~\cite{Warren2018} 
exploited previous \vtr{} experiences on a drone to actively
control a gimbal system so as to reduce the orientation error between the
camera and the viewpoint used while recording experiences.

Recently, Zhang~\etal~presented a perception-aware navigation approach, named
\emph{Fisher Information Fields}~\cite{Zhang2018a, Zhang2019a}. This 
map representation allocates the expected localization performance (given by 
the Fisher Information matrix) in a discrete grid, which is used to compute the 
expected information for arbitrary poses. While we base our localization performance metrics on 
similar principles, in this work we focus on topo-metric representations
instead.

\section{Method}
\label{sec:method}
Our goal is to develop a multi-camera Visual Teach and Repeat system for 
robots with multiple cameras, with a focus on quadruped robots.

\begin{figure}[t]
	\vspace{2mm}
	\centering
	\includegraphics[width=0.48\textwidth]{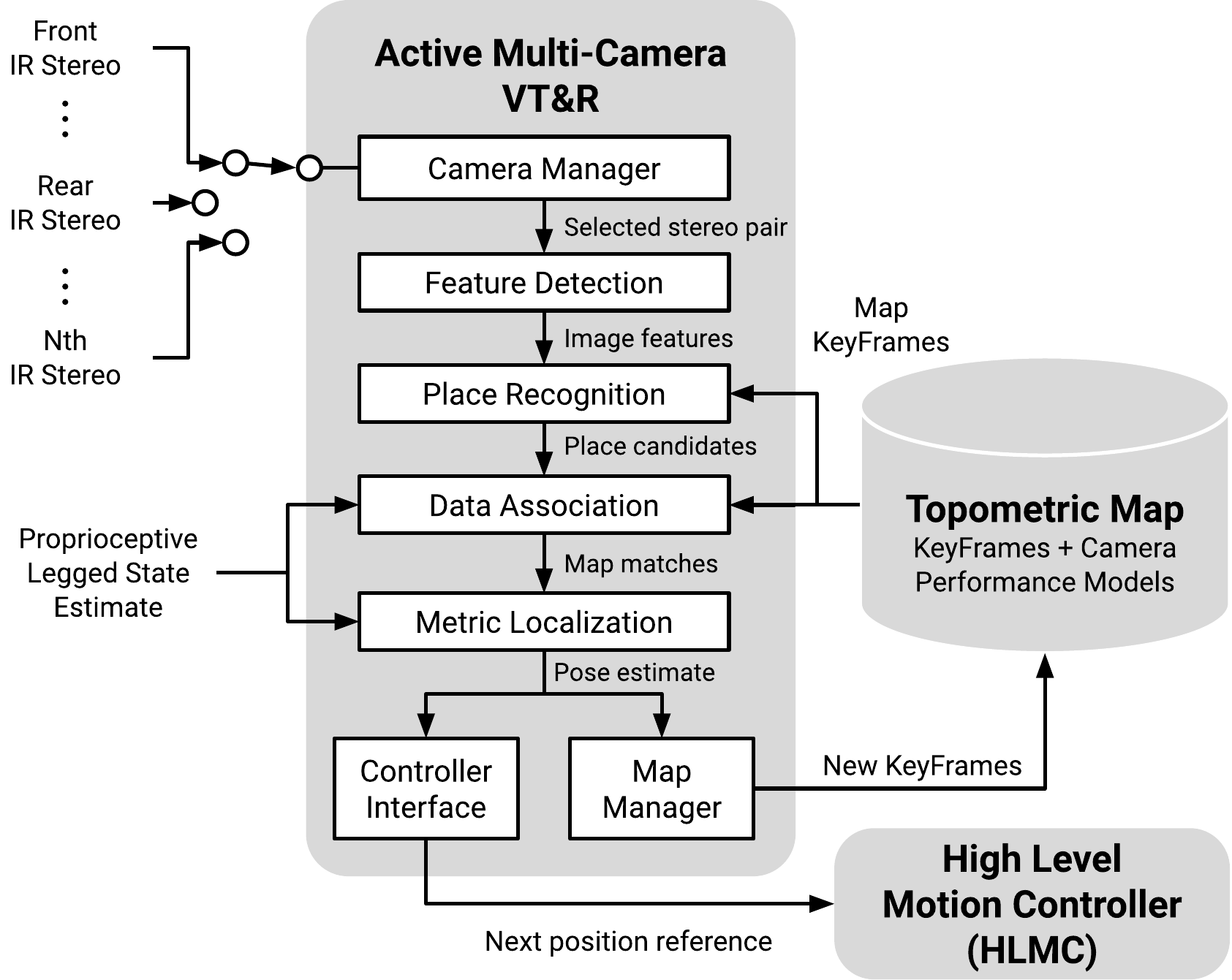}
	\caption{\small{Block diagram of our active multi-camera \vtr{} system. We
			augmented the topo-metric map with a Camera Performance 
			Model (CPM) for each camera. The Camera Manager queries the models
online to 	select
			the active camera.}}
	\label{fig:system-overview}
	\vspace{-5mm}
\end{figure}

\subsection{System Overview}
\label{subsec:system-overview}
The main modules of our system are shown in \figref{fig:system-overview}. The 
structure follows the approach taken by Furgale and 
Barfoot~\cite{Furgale2010}, with three main differences:
\begin{itemize}
	\item We use a slowly drifting proprioceptive state estimate (for a legged 
	robot, provided by a system such as TSIF~\cite{Bloesch2017a})
	instead of visual odometry to simplify the mapping process and 
	reduce the computational burden in the teach step 
	(\secref{subsec:topo-metric-mapping}).
	\item The topo-metric map is augmented with a \emph{Camera Performance 
	Model} (CPM) for each camera, which is learned using the teach
	trajectory (\secref{subsec:learning-entropy-models}).
	\item A new \emph{Camera Manager} module determines which camera should be
	processed at the current instance. During the teach, the manager processes
  every camera in turn. During repeat, it exploits the CPM to select
  the most suitable camera for a specific part of the path 
  (\secref{subsec:repeat-step}).
\end{itemize}

\begin{figure}[t]
	\vspace{2mm}
	\centering
	\includegraphics[width=0.3\textwidth]{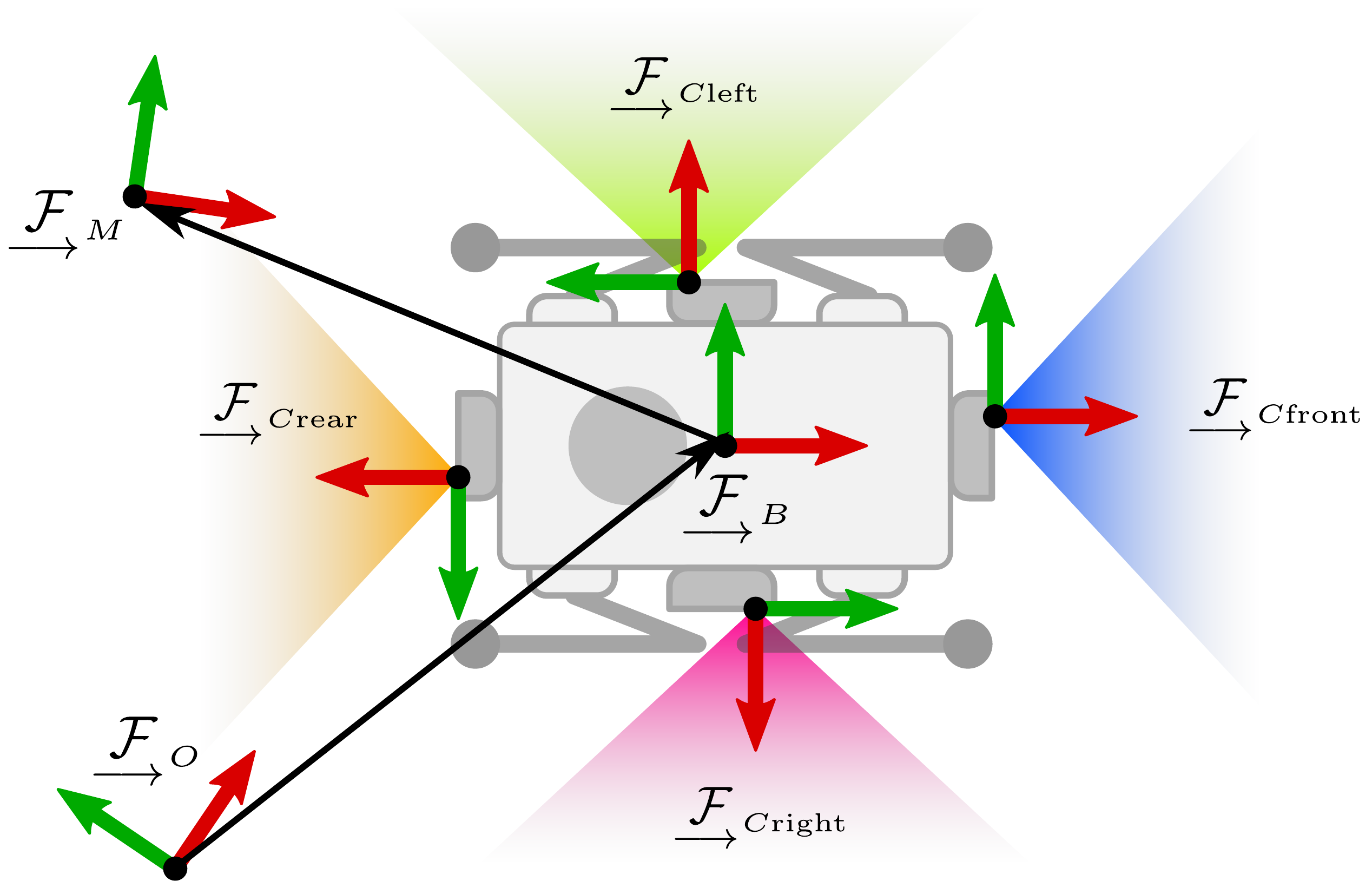}
	\includegraphics[width=0.17\textwidth]{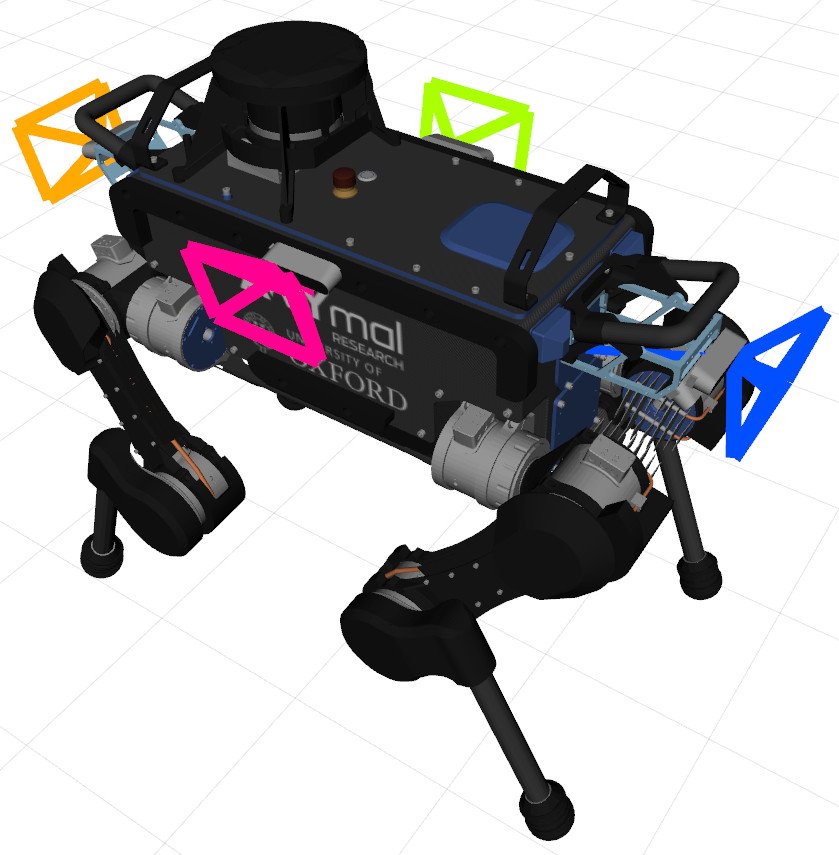}
	\caption{\small{Top-view diagram of the frames and color convention to 
			identify the cameras throughout this paper.}}
    \vspace{-4mm}
	\label{fig:body-frames}
	\vspace{-3mm}
\end{figure}

We consider that up to 4 cameras could be
attached to the main body (front, rear, left and right). 
\figref{fig:body-frames} illustrates the coordinate frames and 
the position of the cameras.

The proprioceptive state estimate is defined in the fixed odometry frame 
$\notatFrame{O}$, while the \vtr{} localization is defined in the fixed map 
frame $\notatFrame{M}$ corresponding to the teach path. The moving frame 
$\notatFrame{B}$ is rigidly attached to the robot's chassis, as well as the 
four camera frames $\notatFrame{C}$.

\begin{figure*}[t]
	\vspace{1mm}
	\centering
	\includegraphics[width=\textwidth]{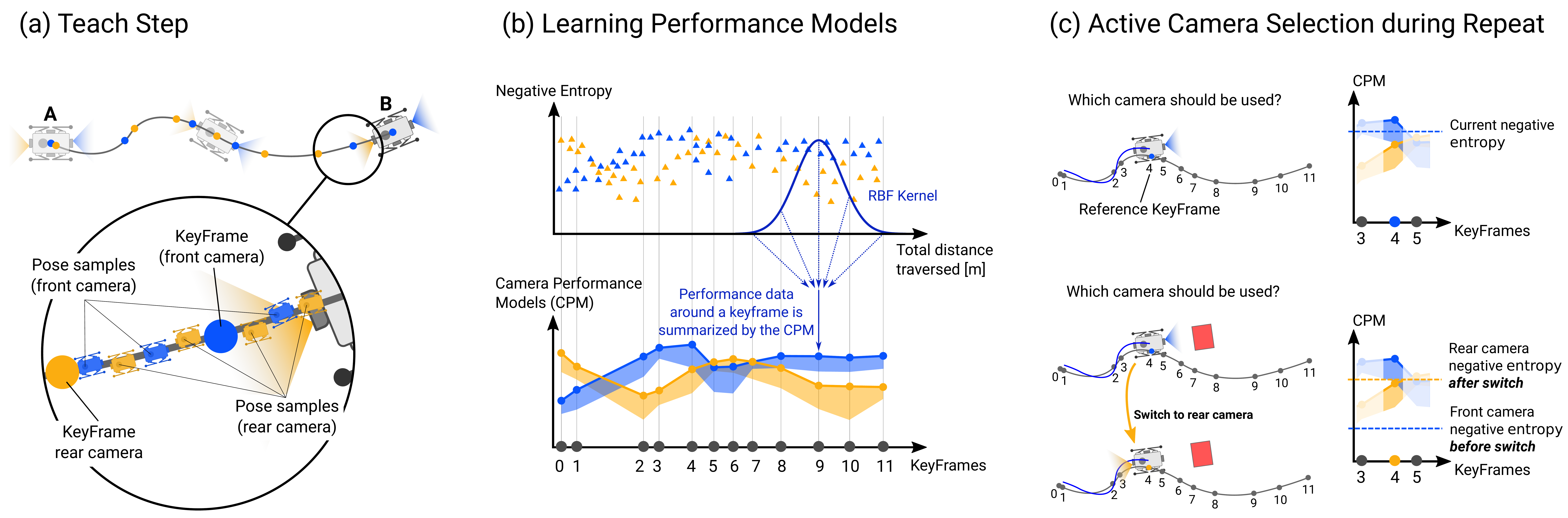}
	\caption{\small{Main steps of our \vtr{} system. \textbf{(a) Teach step:} 
	The	robot is first teleoperated from A to B to build a topo-metric map of 
	the environment. 
	Keyframes are created for each camera, denoted by the different 
	colors. Along with the map, sampled poses and negative
	entropies are also computed and stored.
	\textbf{(b) Learning Performance Models:} For each keyframe, the closest 
	negative entropy samples within a radius $\notationScalar{d}{\text{max}}$ 
	are grouped and are averaged using an RBF kernel to learn a model of
	performance for each camera (\algref{alg:learning-performance}).
	(c) \textbf{Repeat step with active camera selection: } The learned 
	CPMs are used to select the camera with highest predicted performance at
	each segment of the map, or to change the selected camera if the predicted performance 
	is not as expected.}}
	\label{fig:teach-learn-repeat}
    \vspace{-4mm}
\end{figure*}

\subsection{Topo-metric Mapping with Multiple Cameras}
\label{subsec:topo-metric-mapping}
In our system, the teach step builds a topo-metric map of the path over which 
the robot is teleoperated. The map is represented by a collection of
\emph{keyframes} $\notatManifold{K}$ connected by relative transformations (as
in \cite{Churchill2012a}) which are expressed in the map frame
$\notatFrame{M}$ relative to the body frame $\notatFrame{B}$. Each keyframe 
$k\in\notatManifold{K}$ stores:
\begin{itemize}
	\item A stereo image pair, tagged with the source camera.
	\item Triangulated AKAZE~\cite{Alcantarilla2013a} features 
	$\notatManifold{P}$ for metric pose estimation relative to the 
	keyframe.
	\item A Bag-of-Visual-Words vector, based on 
	DBoW2~\cite{Galvez-Lopez2012}.
	\item The body pose of the keyframe in the map frame
	$\notationMatrix{T}{MB} \in \SEthree$. 
	\item The intrinsic calibration between the camera and the body pose 
	$\notationMatrix{T}{BC} \in \SEthree$.
	\item A CPM of each available camera in the current setup.
\end{itemize}

The teach step performs the mapping process using a single camera at a time. 
For each stereo image frame, the \emph{Feature Detector} module extracts 
features $\notatManifold{Z}$, which are then matched against 
the current map by the \emph{Data Association} module. We use the quadruped's 
proprioceptive state estimate as a motion prior which helps to guide 
feature matching.

The \emph{Metric Localization} module uses the matches to 
perform a registration against the map points 
$\notatManifold{P}$ in the last created keyframe: we first use 
Perspective-n-Points (PnP) to obtain an initial estimate, which is later 
refined via pose-only optimization using the reprojection residual 
$\notationVector{r}{\text{reproj}}(\notatVector{z},\notatVector{p})$ with 
covariance $\notationMatrix{\Sigma}{\text{reproj}}$, and a pose prior 
residual $\notationVector{r}{\text{prior}}$ given by the previous estimate 
$\notationMatrix{T}{\text{prior}}$ with covariance 
$\notationMatrix{\Sigma}{\text{prior}}$:
\begin{equation}
\label{eq:pose-optimization}
\argmin
\sum_{\notatVector{z}\in\notatManifold{Z}, \notatVector{p}\in\notatManifold{P}} 
\mahalanobisNorm{\notationVector{r}{\text{reproj}}(\notatVector{z},\notatVector{p})}
 + 
\mahalanobisNorm{\notationVector{r}{\text{prior}}(\notationMatrix{T}{\text{prior}})}{\notationMatrix{\Sigma}{\text{prior}}}
\end{equation}

From this optimization we recover an 
estimate of the covariance $\notationMatrix{\Sigma}{\text{visual}}$ of the 
optimization solution $\notationMatrix{T}{\text{visual}}$, given by 
the Fisher information matrix \cite{Kuo2020}.
Further, the covariance is used to compute the negative entropy 
$\notatScalar{E}$:
\begin{equation}
\notatScalar{E} = -\log( \left|\notationMatrix{\Sigma}{\text{visual}} \right|)
\end{equation}
$\notatScalar{E}$ is a scalar that characterizes the performance of the 
localization at a given pose: a larger negative entropy implies a better 
localization estimate, and vice-versa. An advantage of this method over other 
criteria, such as the number of tracked features, is that it characterizes
the whole localization process. For instance, tracking a small number of
nearby features or a large number of distant features are treated similarly,
because both situations lead to poor localization estimates.
$\notatScalar{E}$ is of particular importance for our system since:

\begin{itemize}
	\item It is used by the \emph{Map Manager} module as a criterion when 
	creating new keyframes, using the running average filter strategy proposed 
	by Kuo 	\etal~\cite{Kuo2020}.
	\item The poses and negative entropies of frames that are not used to 
	create new keyframes are stored in the neighbor keyframes as 
	\emph{performance samples} 	$\notatManifold{S}$ of the actual camera in a 
	specific part of the path.
\end{itemize}

The process is executed for each camera individually, so as to sample their 
performance assuming no other cameras are available. This could 
generate inconsistent trajectories for each camera, so we enforce smoothness 
along the path by prioritizing the use of the proprioceptive state estimate for 
the teach step. The output is shown in \figref{fig:teach-learn-repeat} (a).

\subsection{Learning Performance Models for Each Camera}
\label{subsec:learning-entropy-models}
As previously described, the teach step generates not only a 
topo-metric map, but also a set of performance samples $\notatManifold{S}$ 
from the whole path. A performance sample $s$ is defined as a tuple
$(\notationMatrix{T}{s}, \notationScalar{E}{s}, \notationScalar{c}{s})$, where 
$\notationMatrix{T}{s}$ is the pose of the sample expressed in the map frame, 
$\notationScalar{E}{s}$ is the negative entropy computed for that specific 
pose, and $\notationScalar{c}{s}$ is a tag to identify the particular camera 
that generated that sample.

We use the samples to learn a CPM, which is a model that embeds the performance 
of a camera for a given teach path. The CPM for a single camera $c$ is 
expressed by a collection of Gaussian distributions defined for 
every keyframe in the path. Their parameters $(\notationScalar{\mu}{c}, 
\notationScalar{\sigma}{c})$ are the result of the learning process. 
We define CPMs for all the cameras available in the robot, and we are able to 
query them at each keyframe of the path to determine which camera is 
likely to provide the best localization estimate.

The learning algorithm, defined in \algref{alg:learning-performance},
performs a spatially weighted averaging of samples around each 
keyframe. The averaging weights are computed using a \emph{radial basis 
function} kernel~\cite{Bishop2006} denoted by
$\notatScalar{\kappa}(\notationMatrix{T}{1}, \notationMatrix{T}{2})$ and defined as:
\begin{equation}
\label{eq:kernel}
\notatScalar{\kappa}(\notationMatrix{T}{1}, \notationMatrix{T}{2}) = 
\exp\left(- \frac{ d(\notationMatrix{T}{1}, \notationMatrix{T}{2})^{2} }{2 l}\right)
\end{equation}
where, $d(\cdot, \cdot)$ is a function that computes the distance between the
two poses $\notationMatrix{T}{1}$ and $\notationMatrix{T}{2}$, ignoring the 
rotational component, while $l$ is a hyperparameter 
(scale length) that controls smoothness. \figref{fig:teach-learn-repeat} (b)
shows the output of the learning process for all the keyframes along a given path. 
This is similar to Gaussian Process regression, but is defined on the
discrete space of keyframes, independently from spatial coordinates.

\begin{algorithm}[t]
	\caption{\small{CPM learning using teach path}}
	\label{alg:learning-performance}
	\vspace{2pt}
	\KwIn{Keyframes $\notatManifold{K}$, performance samples 
		$\notatManifold{S}$, maximum distance 
		for sample search $\notationScalar{d}{\text{max}}$, kernel weighting 
		function $\notatScalar{\kappa}(\cdot, \cdot)$}
	\KwOut{CPM for each camera}
	
	\ForEach{keyframe $k$ in $\notatManifold{K}$}{
		$\notationMatrix{T}{k} \leftarrow $ GetKeyFramePose($k$)
		
		$\notatManifold{S*} \leftarrow $ SearchSamplesWithinRadius($k$, 
		$\notatManifold{S}$, 	
		$\notationScalar{d}{\text{max}}$)
		
		\ForEach{camera $c$ in $\notatManifold{S*}$}{
			$ \displaystyle \notationScalar{w}{c} \leftarrow \sum_{s \in 
				\notatManifold{S*}} \notatScalar{\kappa}(\notationMatrix{T}{s}, 
			\notationMatrix{T}{k})  $
			\vspace{0.3\baselineskip}
			
			$ \displaystyle \notationScalar{\mu}{c} \leftarrow 
			\frac{1}{\notationScalar{w}{c}} \sum_{s \in 
				\notatManifold{S*}} \notationScalar{E}{s} \ 
			\notatScalar{\kappa}(\notationMatrix{T}{s}, 
			\notationMatrix{T}{k})  $
			\vspace{0.3\baselineskip}
			
			$ \displaystyle \notationScalar{\sigma}{c} \leftarrow \sqrt{ 
				\frac{1}{\notationScalar{w}{c}} 
				\sum_{s 
					\in 
					\notatManifold{S*}} (\notationScalar{E}{s} - 
				\notationScalar{\mu}{c})^{2} \ 
				\notatScalar{\kappa}(\notationMatrix{T}{s}, 
				\notationMatrix{T}{k}) } $ 
			\vspace{0.3\baselineskip}
			
			UpdateCPM($k$, $c$, $\notationScalar{\mu}{c}$, 
			$\notationScalar{\sigma}{c}$)
		}
	}
\end{algorithm}

\subsection{Repeat Step with Active Camera Selection}
\label{subsec:repeat-step}
The repeat step involves different procedures depending on the status of the 
system:

\paragraph{Global Relocalization} If the system is initializing, the status is
\emph{lost}, and an arbitrary camera will be chosen to attempt 
relocalization. First, the Place Recognition module searches candidate 
keyframes using Bag-of-Words. Then, the Data Association module performs a
standard descriptor matching. The matches are later verified by the
Metric Localization module by attempting a PnP registration and then using 
optimization refinement to discard outliers. The keyframe with the most matches 
is selected as the reference keyframe and the system status is set to
\emph{localized}.
Since the previous procedure is agnostic to which camera generated the 
keyframe, we also compute the reference keyframes for the other available
cameras. This is done by checking all the neighbor keyframes and associating to 
each camera the closest keyframe with the most similar orientation.

\paragraph{Path Traversal with Active Camera Selection} Once the system
has computed the reference keyframes and an initial pose has been estimated, the
repeat step is ready for execution. Details regarding the integration with the
quadruped's controller are described later in 
\secref{subsec:quadruped-integration}.

While the robot traverses the path and the Feature Detection, Data Association,
and Metric Localization are being executed for the teach step, the
Camera Manager module analyzes the different image 
streams and \emph{actively} changes the current camera if: 1) there is
another camera that can provide better performance at that specific part of 
the path, or 2) the current performance is not as the model describes,
typically due to a change in the environment.

The first case only requires to query each 
$\notationScalar{\mu}{c}$ in the CPM to find the best camera for the current 
reference keyframe. The second case requires comparison between the current negative entropy 
$\notationScalar{E}{t}$ and a lower bound $\notationScalar{E}{t} < 
\notationScalar{\mu}{c} - \notatScalar{k} 
\notationScalar{\sigma}{c}$ computed from the CPM parameters 
$(\notationScalar{\mu}{c}, \notationScalar{\sigma}{c})$
associated to the reference keyframe. The bound defines a \emph{margin} so as 
to not select a new camera unless performance has decreased 
significantly, which can be tuned with the hyperparameter $k$.

In general, with accurate path tracking and with similar visual conditions in 
the teach and repeat steps, the previous inequality will never be satisfied, 
and the negative entropy will stay within the limits. However, if the 
environment changes, the feature extraction will be affected, potentially 
degrading the visual pose estimate and decreasing the negative entropy below 
the lower bound.
When this occurs, the camera is flagged and cannot be used for a fixed time.
The remaining CPMs are queried to find the next best camera for the given 
path section. An example of this procedure is illustrated in 
\figref{fig:teach-learn-repeat} (c).
If all the cameras are flagged and the system loses visual 
tracking, it reports \emph{Tracking Lost}.

\paragraph{Tracking Lost} When the system loses visual tracking, we use the 
last successful localization estimate and predict the current pose by using 
relative motion estimates from the legged proprioceptive state estimator.
Meanwhile, the system will attempt a re-localization by matching against
the neighbor keyframes within a certain radius, a procedure we call
\emph{Local Relocalization}. If the system cannot succeed after \SI{10} seconds,
it declares itself \emph{lost} and will stop navigating until it is reset by
the user.

\subsection{Closed-loop Integration in a Legged Robot}
\label{subsec:quadruped-integration}
The motion of a wheeled robot or a car is smooth and without any sharp jerks. A
state-of-the-art \vtr{} system can track precisely enough to keep a UGV within the tram-lines 
of previous runs (as in \cite{Furgale2010}). This degree of smoothness has a
 stabilizing effect on \vtr{} localization. In contrast, the same degree of 
 smoothness in
camera motion is not possible on legged robots. The robot's gait
induces a sharp, jerky motion, so that exact teach
trajectory tracking is impossible.

The quadruped's Whole Body Controller (WBC) controls its 12 joints to achieve
goals such as a desired base velocity. We interface with it through a High Level Motion Controller
(HLMC), which computes a base velocity reference given a desired base pose. The \vtr{} system
generates a sequence of waypoints from the teach path expressed in the map
frame. Given the current localization estimate, the \vtr{} system selects the closest
waypoint to the robot and sends it as the next desired base pose reference to the HLMC. In
this way, we circumvent the need to precisely replicate the base motions of
original teach trajectory, but we keep the robot close to it at all times.

Finally, in contrast to wheeled platforms, legged robots are holonomic and can
strafe or turn in place to execute inspection tasks; we illustrate how the
\vtr{} handles such situations in our attached video.

\section{Experimental Results}
\label{sec:experiments}
Experiments were performed with an ANYmal B300 robot equipped with two
unsynchronized Intel RealSense D435i stereo cameras angled down by 
\SI{12}{\degree}; we used the IR stereo pairs as the visual input for our 
system. The robot also carries a Velodyne VLP-16 LiDAR, which we
used in post-processing to obtain \SI{10}{\hertz} ground truth by registering 
scans within a prior map using Iterative Closest Point (ICP)~\cite{Besl1992}. 
Our system ran onboard on a single Intel i5 CPU shared along with other required
modules and drivers.

For evaluation, we compared repeat trajectories to the initial teach run. We
determined the instantaneous tracking error as the perpendicular distance
between each pose during the repeat run and a line fit to the
nearest neighbor points of the teach step path. The mean \emph{Path Tracking 
Error} (PTE) is a measure of tracking performance for a full run.

Our \vtr{} system was tested in two experimental scenarios: an indoor
workshop (E1) and a larger industrial environment outdoor (E2).
\tabref{tab:pte} summarizes the path tracking performance for all the runs
using the \vtr{} pose estimates. Finally, we ran our system in simulation 
with 4 cameras to demonstrate our approach with more complex camera setups.

\begin{figure}[t]
	\vspace{3mm}
	\centering
	\includegraphics[width=0.9\columnwidth]{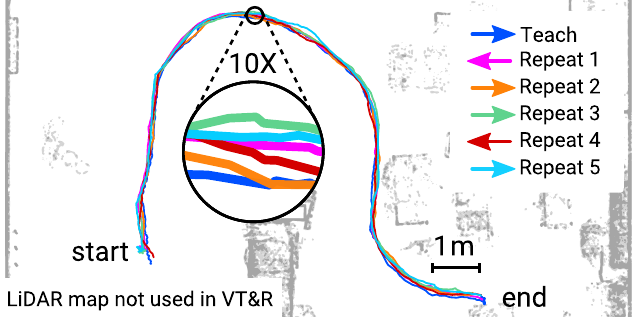}
	\caption{\small{Experiment 1 (Indoor): Ground truth trajectories of teach 
	(blue) and repeat runs. Direction of motion is indicated by the legend. 
	Overall, the robot never exceeded \SI{20}{\centi\meter} of path
	tracking error, with an average of \SI{7}{\centi\meter}.}}
	\label{fig:indoor}
	\vspace{-2mm}
\end{figure}

\begin{table}[t]
	\vspace{3mm}
	\centering
	\caption{\small{Quantitative results for the indoor (E1, 5
			repeats) and outdoor experiments (E2, 4 repeats).}}
	\begin{tabularx}{\columnwidth}{cc*{5}{Y}}
		\thickhline
		\multicolumn{6}{c}{\textbf{Path Tracking Error (PTE)}
		$\boldsymbol{\mu}\pm \boldsymbol{\sigma}$ 
			[\si{\metre}]} \\
	    \hline
		& \textbf{R1} & \textbf{R2} & \textbf{R3} & \textbf{R4} & \textbf{R5} \\
		\hline
		\multicolumn{1}{l|}{\textbf{E1}} &0.05~\textpm~0.03 & 0.06~\textpm~0.03 
		& 
		0.09~\textpm~0.06 & 0.06~\textpm~0.03 & 0.09~\textpm~0.04\\ \hline
		\multicolumn{1}{l|}{\textbf{E2}} & 0.09~\textpm~0.04 & 
		0.13~\textpm~0.05 & 
		0.07~\textpm~0.05 & 0.14~\textpm~0.07 & -\\ 
		\thickhline
	\end{tabularx}
	
	\label{tab:pte}
\end{table}


\begin{figure}[t!]
	\centering
	\includegraphics[width=\columnwidth]{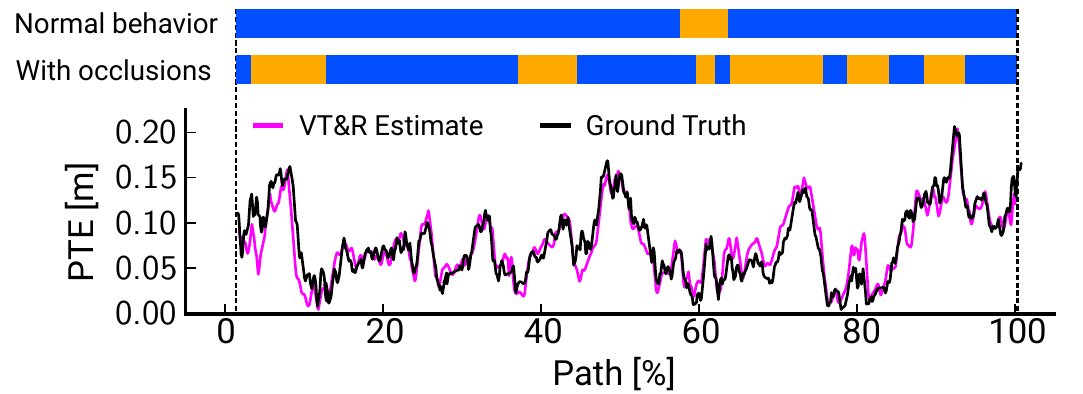}
	\caption{\small{Indoor experiment (E1): Estimated (magenta) and ground truth
	(black) PTE between teach and repeat R5. Color bars show the switch between the front
camera (blue) and the rear camera (orange) in normal operation (top) and when occluded
(bottom).}}
	\label{fig:errors}
	\vspace{-4mm}
\end{figure}

\begin{figure*}[t]
	\vspace{3mm}
	\centering
	\includegraphics[width=0.9\textwidth]{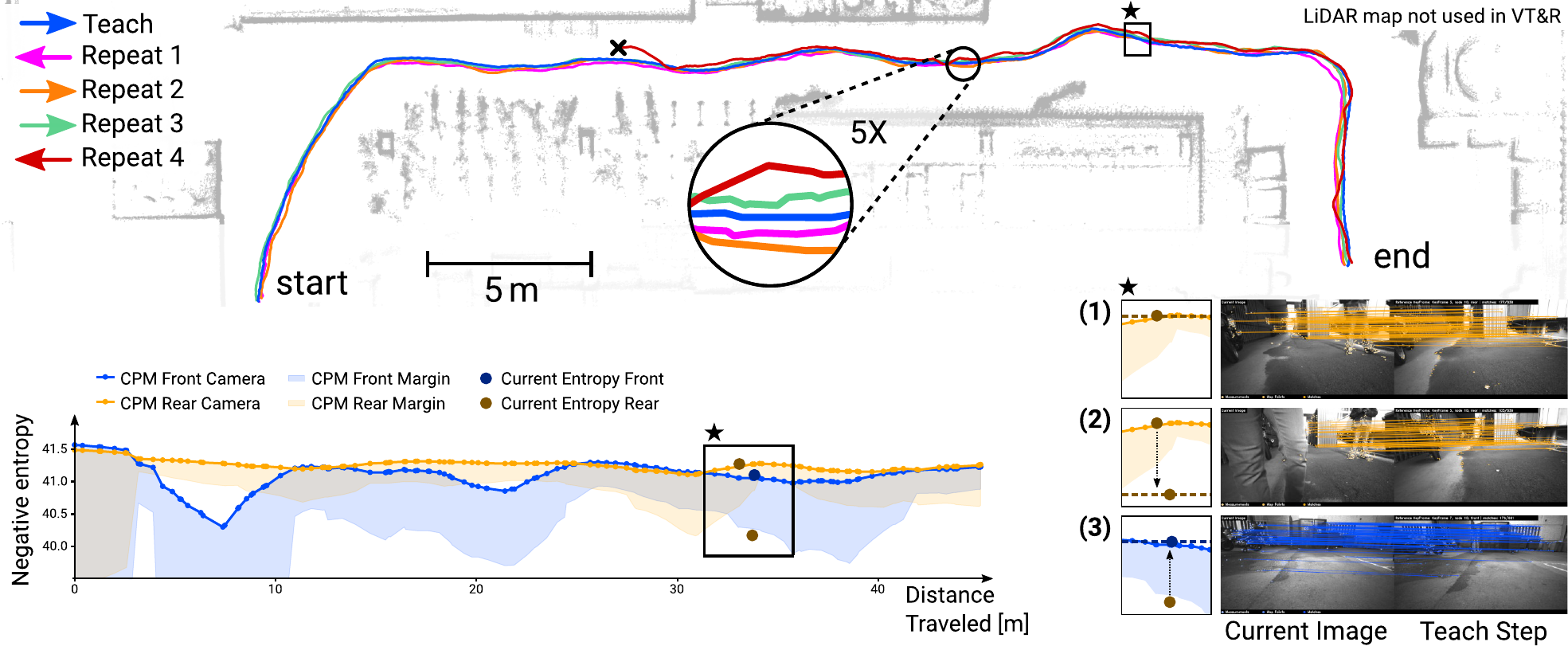}
	\caption{\small{Outdoor experiment (E2): \emph{Top:} Ground truth
	trajectories of the teach (blue) and repeat runs. The robot never
	exceeded \SI{20}{\centi\meter} of tracking error, with an average of
	\SI{11}{\centi\meter}. \emph{Bottom:} Left plot illustrates the CPM
	computed for the whole path, right images are examples of matches between 
	the live stream and the teach path. 
	A segment of the trajectory during Repeat 2 in which the
	camera is occluded is marked with a $\mathbf{\bigstar}$ symbol:
	(1) Rear camera performance (in orange) is within the CPM limits.
	(2) The camera is occluded, leading to a drop in the negative entropy,
	triggering a camera switch. 
	(3) After the system switched to the front camera, its current negative	
	entropy (blue) is closer to the CPM prediction.}}
	\label{fig:outdoor}
	\vspace{-2mm}
\end{figure*}

\subsection{Experiment 1: Indoor}
We first performed a series of experiments in a cluttered lab environment. The
robot was teleoperated to walk between furniture, machines and other equipment,
covering a distance of \SI{15}{\meter}. It then autonomously returned to the
initial position (backwards), and repeated the path back and forth 5 times. 

The robot demonstrated stable navigation in all these runs, and it was able to 
stay within \SI{20}{\centi\meter} of the teach path at all times, regardless of 
the walking direction (\figref{fig:indoor}). Numerical 
comparisons in \tabref{tab:pte}, demonstrate the low tracking error obtained 
for each run.

In \figref{fig:errors}, we evaluated the tracking performance (as estimated by 
\vtr{} system online) by comparing it to the true tracking error (computed 
using the LiDAR ground truth) for the fifth repeat run of these experiments. 
The high degrees of correlation between the two estimates demonstrates that the 
\vtr{} system can accurately localize the robot against the teach path even if 
one of the cameras is occluded. Deviations are due to the shape of the teach 
path and the responsiveness of the tracking controller.

\subsection{Experiment 2: Outdoor}
For our second experiment, we tested our system in an outdoor environment
with adverse lighting and repetitive, industrial structure. We
teleoperated the robot to walk a \SI{45}{meter} long path, which was 
successfully traversed 3 times in repeat mode (\figref{fig:outdoor}). On a 
fourth repeat run the system was interrupted after the localization module 
diverged due to poor visual feature tracking. It was caused by changing 
lighting conditions, which is subject to future work.

During the second repeat run we occluded the cameras by having a person walking 
in front of the robot. Our \vtr{} successfully changed the active camera and 
completed the mission regardless. If the robot had used a single camera (i.e, 
no active selection available), such situations could have severely affected 
the visual tracking, degrading the performance or even failing to finish the 
mission in case of prolonged occlusions.

\subsection{Experiment 3: Qualitative Experiments with 4 Cameras}
\begin{figure}[t]
	\centering
	\includegraphics[width=0.85\columnwidth]{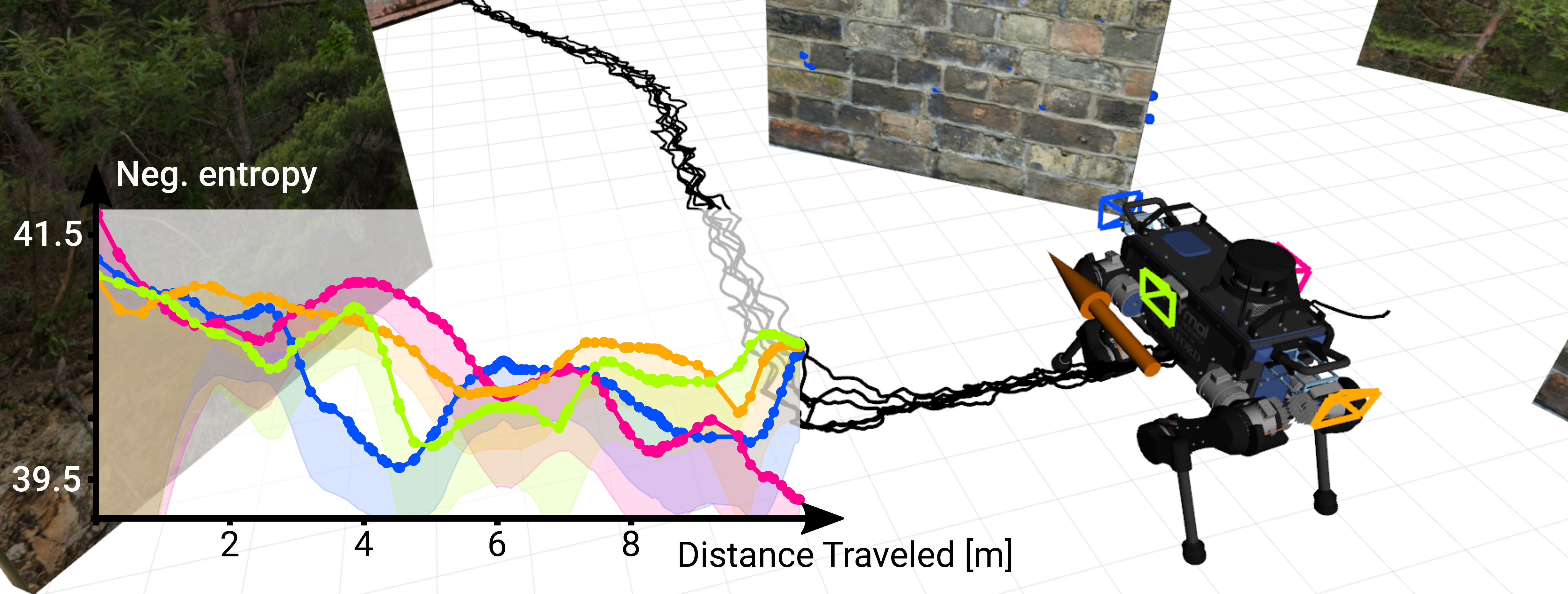}
	\caption{\small{Experiment 3 (Simulation):
	Tracking performance and CPM with 4 cameras. Black lines
	denote the ground truth paths for 6 consecutive repeats. The
	orange arrow is the next waypoint.
	}}
	\label{fig:simulation}
	\vspace{-3mm}
\end{figure}

Lastly, we performed experiments in simulation by equipping the ANYmal with 4 
cameras. The goal was to demonstrate that our approach naturally generalizes to 
other camera configurations and is applicable to the latest version of
ANYmal, the C-series, which has a similar 4 camera configuration.
For the teach step, we made the robot walk through the environment with 
motion in all directions (forward, sideways and turning). \figref{fig:simulation} shows an example 
of the trajectories traversed in the simulation. Further experiments with the 
real robot will be a focus of future work.

\section{Conclusions}
\label{sec:conclusions}
We presented a novel \vtr{} system that utilizes multiple
non-synchronized cameras to perform autonomous point-to-point navigation. By 
exploiting information collected during a teach step, we learned a performance
model for each camera that preserved the topo-metric structure. We
demonstrated how the system utilized the learned models online to actively 
select the most informative camera and be resilient to sudden changes in the
environment due to occlusions.

In a series of real and simulated navigation scenarios on a quadruped robot,
our system successfully followed a previously taught route in spite of the 
complexities of jerky motion and people (intentionally) occluding
the cameras.

In future, we plan to extend our \vtr{} system with other visual cues to
improve its performance in more complex locomotion regimes such as stair
climbing and obstacle traversals which cause the visual scene to change
more dramatically.


\section*{Acknowledgement}
This research is supported by
the ESPRC/UKRI ORCA Robotics Hub
(EP/R026173/1), a Royal Society University Research Fellowship (Fallon)
and the National Agency for Research and 
Development of Chile (ANID)~/~Scholarship Program~/~DOCTORADO BECAS 
CHILE~/~2019-72200291 (Mattamala).


\balance
\bibliographystyle{./IEEEtran}
\bibliography{./IEEEabrv,./library,../library}

\end{document}